# Beyond Specialization: Benchmarking LLMs for Transliteration of Indian Languages

Gulfarogh Azam, Mohd Sadique, Saif Ali, Mohammad Nadeem, Erik Cambria, Shahab Saquib Sohail, Mohammad Sultan Alam

**Abstract**—Transliteration, the process of mapping text from one script to another, plays a crucial role in multilingual natural language processing, especially within linguistically diverse contexts such as India. Despite significant advancements through specialized models like IndicXlit, recent developments in large language models (LLMs) suggest a potential for general-purpose models to excel at this task without explicit task-specific training. The current work systematically evaluates the performance of prominent LLMs—including GPT-4o, GPT-4.5, GPT-4.1, Gemma-3-27B-it, and Mistral-Large against IndicXlit, a state-of-the-art transliteration model, across ten major Indian languages. Experiments utilized standard benchmarks, including Dakshina and Aksharantar datasets (AK-Freq, AK-NEF, AK-NEI), with performance assessed via Top-1 Accuracy and Character Error Rate (CER). Our findings reveal that while GPT family models generally outperform other LLMs and IndicXlit for most instances. Additionally, fine-tuning GPT-4o improves performance on specific languages notably. An extensive error analysis and robustness testing under noisy conditions further elucidate strengths of LLMs compared to specialized models, highlighting the efficacy of foundational models (or appropriately tuned LLMs) for a wide spectrum of specialized applications with minimal overhead.

✦

## 1 Introduction

INDIA'S linguistic landscape is remarkably diverse, with 22 officially recognized languages spanning multiple language families and writing systems [1], [2]. Many of Indian languages use distinct native scripts (e.g., Devanagari for Hindi, Tamil script for Tamil, Perso-Arabic script for Urdu), yet it is common for speakers to romanize text – writing their languages using the Latin (English) script – especially in informal digital communication [3], [4]. Such prevalence of cross-script text and the need to bridge multiple scripts make transliteration a crucial task in Indian NLP [5]. Effective transliteration enables information retrieval and processing across scripts. For example, converting names or queries between native and Latin scripts is often necessary in machine translation, cross-language information retrieval, and knowledge integration [6]. Therefore, developing accurate methods to transliterate Romanized script to native Indian languages (and vice versa) is vital for both usability and interoperability in multilingual applications.

Traditionally, transliteration literature focused primarily on proper names and isolated words [6], [7]. Finite-state models and phrase-based systems were popular in the past, but sequence-to-sequence neural models have since become dominant [8], [9] and yield higher accuracy in mapping between scripts [10]. In the context of Indic languages, various parallel corpora and datasets have been created to support transliteration research [3]. Notable among these is the Dakshina dataset [4] which provides a collection of native-script text, romanized lexicons, and parallel sentences for 12 South Asian languages. Dakshina enabled baseline systems for single-word and sentence-level transliteration, but its scale (around 300k word pairs) is limited relative to the diversity of Indic languages. Recently, the Aksharantar project vastly expanded available data, releasing a 26-million example transliteration corpus covering 21 Indian languages [3]. Accompanied by a 103k-word evaluation set partitioned by word frequency and origin (native vs. foreign). The authors of [3] also developed IndicXlit model (a transformer-based multilingual transliterator) which was trained on Aksharantar dataset and achieved state-of-the-art results, with a 15% accuracy improvement on the Dakshina benchmark [3].

Recently, the NLP field has been revolutionized by the emergence of large language models (LLMs) that are general-purpose [11], [12]. Models such as GPT-3 and GPT-4 have shown an ability to perform a wide range of tasks in a zero-shot or few-shot manner after being trained on massive multilingual data [13]. These foundation models encode cross-lingual knowledge, raising the question of whether transliteration can be handled by general LLMs without explicit retraining. Moreover, whether pretrained language models can excel at transliteration after fine-tuning. Kirov et al. [7] showed that fine-tuning large pretrained models yields improvements on sentence-level transliteration for all Dakshina languages. However, it remains unclear how general-purpose LLM (without task-specific fine-tuning) perform when compared against

---

- Mohammad Nadeem is with the Department of Computer Science, Aligarh Muslim University, Aligarh, India.
  E-mail: nadeem.amu@gmail.com, mnadeem.cs@amu.ac.in
- Erik Cambria is with the College of Computing and Data Science, Nanyang Technological University, Singapore.
  E-mail: cambria@ntu.edu.sg
- Shahab Saquib Sohail and Mohammad Sultan Alam is with School of Computing Science and Engineering, VIT Bhopal University, Bhopal-Indore Highway, Kothrikalan, Sehore, Madhya Pradesh, 466114, India.
  E-mail: shahabsaquibsohail@vitbhopal.ac.in

Manuscript received XXXX; revised XXXX, accepted XXXX, published XXXX. This work is funded by ....
Corresponding author: Mohammad Nadeem

a task-specific systems (such as IndicXlit) on transliteration benchmarks. Given that transliteration requires mapping character sequences between scripts, it is an open question whether a general LLM's knowledge is sufficient for high accuracy, especially on out-of-vocabulary or rare words.

In this work, we put forth a comprehensive evaluation of general-purpose LLMs versus a specialized transliteration model for Indian languages. We benchmark five recent LLMs – GPT-4o, its newer variant GPT-4.5, GPT-4.1, Gemma-3-27B-it (a 27B-parameter instruction-tuned model), and Mistral-Large – against the IndicXlit model [3]. Experiments cover ten diverse Indian languages: Bengali, Gujarati, Hindi, Kannada, Malayalam, Marathi, Panjabi, Tamil, Telugu, and Urdu. We evaluate on standard datasets, using the Dakshina test set and the Aksharantar evaluation subsets (specifically, the frequent-word set *AK-Freq* and two named entity subsets *AK-NEF* and *AK-NEI*). Evaluation is conducted with common transliteration metrics: Top-1 Accuracy and Character Error Rate (CER). Our findings shed light on the capabilities and limitations of LLMs for transliteration. In general, the IndicXlit model outperforms most LLMs except GPT-4.5 for most instances. For AK-Freq benchmark, IndicXlit ouperformed GPT-4.5 also in some cases. Overall, GPT-4.5 emerges as the strongest model. We further show that fine-tuning a general model (e.g. GPT-4o) on transliteration data for specific languages can substantially boost its accuracy and it suprpasses all the other models for most scenarios. We also conducted error analysis and robustness test under noisy input conditions (simulated typos and spelling variations) and compared performance of models.

The major contributions of the current study are as follows:

- Conducted a comprehensive comparative analysis of general-purpose LLMs and the specialized IndicXlit model across ten diverse Indian languages using benchmark transliteration datasets.
- Demonstrated that fine-tune improves the performance of general-purpose LLMs for transliteration task.
- Presented error analysis to identify common transliteration errors across different models.
- Evaluated robustness of models through the introduction of input noise.

The rest of the paper is organized as follows. Section 2 discusses the methodology adopted. Section 3 presents the results and discussion of the current work. Limitations of the study are outlined in Section 4. Section 5 presents the concluding remarks and future directions.

## 2 Methodology

The process employed in our methodology work comprises input language data, model inference through different transliteration systems (task-specific, general-purpose, fine-tuned), prediction generation, and evaluation based on metrics like Top-1 Accuracy and Character Error Rate (CER), as shown in Figure 1. Each component of Figure 1 is discussed next.

### 2.1 Datasets

To assess the performance of models, we used two popular and open-source benchmark datasets: the Dakshina dataset and the Aksharantar dataset. Both include rich sets of word pairs with each native script word mapped to its romanized equivalent.

### 2.1.1 Dakshina

Dakshina dataset [4] provides a curated set of high-quality, high-frequency words and sentence pairs from twelve South Asian languages: Bengali, Gujarati, Hindi, Kannada, Malayalam, Marathi, Panjabi, Sindhi, Sinhala, Tamil, Telugu, and Urdu. The languages represent both the Indo-Aryan and Dravidian language families and employ a variety of scripts, ranging from Brahmic to Perso-Arabic. The data was largely drawn from Wikipedia, and its romanized forms were presented and validated by native speakers to ensure accuracy and linguistic authenticity. It contains approximately 300K word pairs and 120K sentence pairs.

### 2.1.2 Aksharantar

Aksharantar dataset [3] is by far the largest publicly available Indic language transliteration dataset. It encompasses more than 26 million word pairs across 21 constitutionally mandated Indian languages, spanning over three different families of Indian languages and covering over 12 scripts. The dataset was collected from several sources, such as Wikidata [14], the **Samanantar** parallel corpus [15], **IndicCorp** monolingual data [16], and manually curated entries. Although Aksharantar is mainly derived from the Dakshina dataset, it considerably increases the data coverage and diversity, making it a more comprehensive benchmark [3]. The Aksharantar dataset is split into training, validation, and test sets. The test set has approximately 103K word pairs for 19 languages and is also divided into four separate subsets:

- **AK-Freq**: It includes high-frequency native words that are the most common vocabulary seen in everyday language use.
- **AK-Uni**: This class of words is equally sampled and chosen to have n-gram diversity to enable the evaluation of generalization over word structures.
- **AK-NEF(Named Entities − Foreign)**: Foreign-origin named entities, including non-Indian names and locations, transliterated into indigenous scripts.
- **AK-NEI(Named Entities − Indian)**: It contains Indian-origin named entities, like Indian names and geographical locations.

For our research work, we chose ten languages that are shared by both the Dakshina and Aksharantar datasets and follow different script styles. The ten languages are Bengali, Gujarati, Hindi, Kannada, Malayalam, Marathi, Panjabi, Tamil, Telugu, and Urdu. In our examination of the Aksharantar test dataset, we noticed that for some of the chosen languages, the AK-Uni subset was missing or incomplete. Therefore, we did not include this category in our assessment to ensure fairness and consistency across all languages. Therefore, for all languages except for Urdu (for Urdu langauge, AK-Freq subset was missing), we used Dakshina test set and AK-Freq, AK-NEF and AK-NEI subsets of Aksharantar test dataset (approximately 123K word pairs). The breakdown of word pairs the final test set for each language is as given in Table 1.

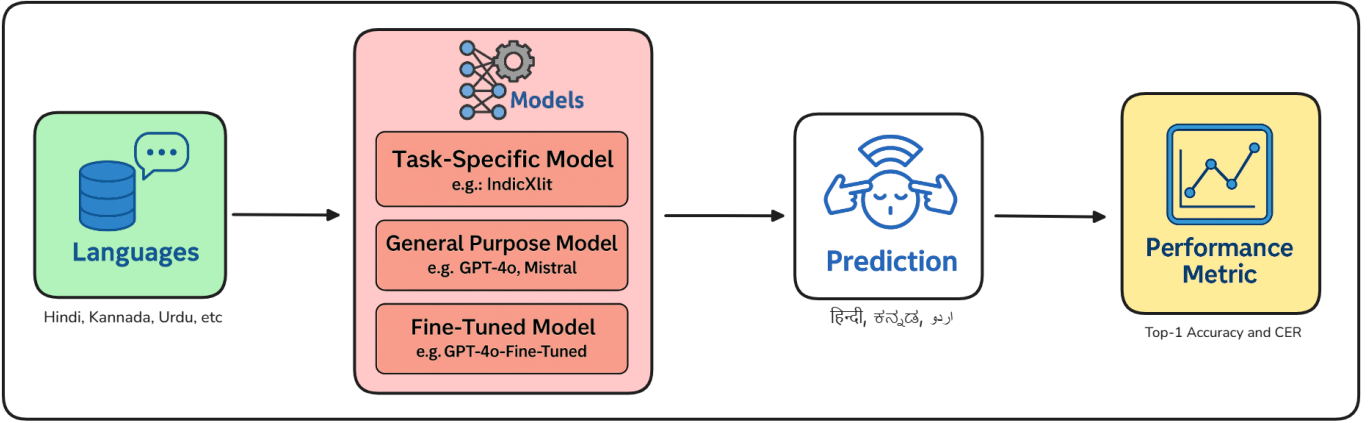

Figure 1: System workflow depicting data preprocessing, model inference, evaluation metrics, and error analysis steps.

| Language | Dakshina | AK-Freq | AK-NEF | AK-NEI | Total |
|---|---|---|---|---|---|
| Bengali | 9198 | 2270 | 1060 | 1639 | 14167 |
| Gujarati | 10343 | 5247 | 1006 | 1481 | 18077 |
| Hindi | 4451 | 3647 | 826 | 1188 | 10112 |
| Kannada | 5051 | 4213 | 878 | 1274 | 11416 |
| Malayalam | 5574 | 4858 | 833 | 1191 | 12456 |
| Marathi | 5645 | 4467 | 831 | 1247 | 12190 |
| Panjabi | 6950 | 2194 | 859 | 1234 | 11237 |
| Tamil | 6847 | 2608 | 829 | 1216 | 11500 |
| Telugu | 5724 | 2241 | 948 | 1347 | 10260 |
| Urdu | 10451 | - | 816 | 1174 | 12441 |
| Total | 70234 | 31745 | 8886 | 12991 | 123856 |

Table 1: Dataset distribution across languages from Dakshina and Aksharantar (AK-Freq, AK-NEF and AK-NEI) test sets

## 2.2 IndicXlit

IndicXlit is a multi-language, multi-script transliteration model, developed by AI4Bharat [3]. It supports 21 languages across 12 scripts from 3 language families, it is one of the state-of-the-art transliteration model for Indian languages. The model is trained on the Aksharantar dataset using transformer based encoder-decoder model [17]. The model works at the character level and is trained in a one-to-many multilingual environment, i.e., a single romanized input can be translated to several Indic scripts based on the target language tag. The input vocabulary is the collection of Roman characters present in the training set (28 Roman characters), and the output vocabulary is the union of characters from different Indic language scripts present in the training set (21 Indic language scripts, summing up to a total of 780 characters). The model has shown superior performance then the previous benchmarks such as the Dakshina dataset, where it records up to 15% better top-1 accuracy [3].

## 2.3 LLMs

Large language models (LLMs) are advanced deep learning models trained on enormous datasets so that they can parse, generate, and manipulate human language with high accuracy and fluency. Advanced LLMs contain billions of parameters and utilize transformer-based networks to capture long-range contextual relationships in text data. Such characteristics make them extremely good at linguistic tasks, such as transliteration. Recent studies have shown advancements in the use of deep learning models for transliteration tasks [18]. In our study, we used several famous general-purpose LLMs as listed next.

### 2.3.1 GPT family

We utilized three OpenAI's GPT models including GPT-4o, GPT-4.5 and GPT-4.1. All used GPT models are autoregressive models. GPT-4o has been trained end-to-end on all different modalities including text, image and audio [19]. GPT-4.5 was trained on a broader, more curated dataset. Initial tests performed on GPT-4.5 reveal that it yields better natural interaction due to its superior knowledge, improved understanding of user needs, and enhanced emotional intelligence [20], [21]. GPT-4.1 has shown good results in instruction-following conditions by scoring 38.3% on Scale's MultiChallenge benchmark [22] surpassing the score of GPT-4o [23].

### 2.3.2 Gemma-3-27b-it

It is among the third generation of releases in the Gemma family of open-source models that Google DeepMind has developed. Gemma 3 presents several architectural innovations with respect to instruction-following and multilingual support in more than 140 languages. Long context handling (up to 128K tokens), efficient memory usage & improved attention mechanisms, and light inference compatibility are its architectural innovations [24], [25].

### 2.3.3 Mistral-Large 2402

It is a major language model from Mistral AI and is mainly designed to support reasoning as well as instruction-following tasks for complex problems [26]. It is a dense, decoder-only transformer model and has 32K token context window. The model also guarantees native fluency in several languages [26], [27].

While these models are not explicitly trained for transliteration tasks, their strong alignment with natural user prompts, multilingualism, and ability to accommodate various inputs rich in context make them promising for transliteration tasks.

## 2.4 Prompt used

To evaluate general purpose LLMs, we tested many prompts. GPT-4o, GPT-4.5, and other OpenAI models use a structured messaging format known as Chat Markup Language (ChatML), in which each input message is tagged with a role (e.g., *system*, *user*, *assistant*) to allow finer control over the model's behavior and contextual understanding [28]. The *system* message defines the model's persona or overall behavior, while the *user* message contains the primary content or instruction. The prompt used in our study is illustrated below.

```
system: You are a skilled transliteration expert,
converting Romanized English to Hindi script.  Your
goal is to produce accurate and readable native Hindi
equivalents.
Example: "input":["namaste","aadar"], "output": ["
नमस्ते","आदर"]
user: Transliterate the given list of Romanized
English words into native Hindi script. Return only a
JSON array.
```

For other models such as Mistral-Large 2402 and Gemma-3-27B-it, a simplified instruction-guided prompt format was used. The format consisted of a instructions followed by a Romanized English input list and an expected transliterated output example.

```
You are a skilled transliteration expert,
converting Romanized English to Hindi script.  Your
goal is to produce accurate and readable native Hindi
equivalents.
Example: "input":["namaste","aadar"], "output":["
नमस्ते","आदर"]    Transliterate the following Romanized
English words into native Hindi script. Return them
as a JSON array.
```

## 2.5 Fine-tuning

In our work, fine-tuning refers to the process of further training a LLM for a specific task, which in our case is transliteration. Prior research has indicated that model performance on specialized tasks could significantly improve with targeted adaptation [29]. We fine-tuned only GPT-4o model as it was the best-performing model among all the fine-tunable models selected and also showed competitive performance against the specialized IndicXlit model.

We fine-tuned GPT-4o using OpenAI API service. The fine-tuning dataset consisted of 982,381 training and 90,140 validation transliteration word pairs, drawn from the Dakshina and Aksharantar datasets. From each of the ten selected Indian languages, approximately 100,000 examples were taken for training. The train and validation dataset distribution can be seen in the Table 2.

The dataset was arranged in the same ChatML format(as mentioned in previous section) required by GPT-4o's (gpt-4o-2024-08-06) fine-tuning pipeline [30]. The process (see Figure 2) followed OpenAI's standard recommendations for formatting and validation [31]. Except learning rate multiplier (value = 1.5), remaining parameters were kept at their default values.

| Language | Training Set | Validation Set |
|---|---|---|
| Bengali | 102843 | 11276 |
| Gujarati | 99223 | 12419 |
| Hindi | 108886 | 6357 |
| Kannada | 99000 | 7025 |
| Malayalam | 97838 | 7613 |
| Marathi | 93384 | 7646 |
| Panjabi | 94276 | 8880 |
| Tamil | 92321 | 8824 |
| Telugu | 102056 | 7681 |
| Urdu | 92554 | 12419 |
| **Total** | **982381** | **90140** |

Table 2: Training and Validation Dataset Statistics across Languages

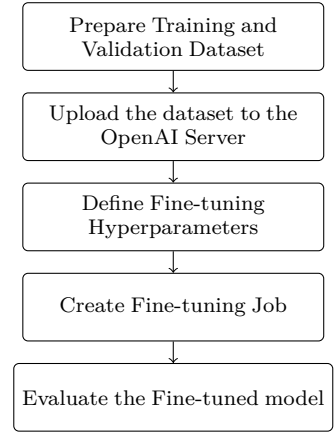

Figure 2: Fine-tuning workflow using OpenAI's API for GPT-4o.

## 2.6 Performance metric

In our study, we have used Top-1 Accuracy and Character Error Rate (CER) to identify the model correctness.

- **Top-1 Accuracy** refers to the percentage of instances where the most probable output (i.e., the model's first prediction) exactly matches the actual data (see Equation (1)).

$$\text{Top-1 Accuracy} = \frac{N_{\text{correct}}}{N_{\text{total}}} \times 100 \quad (1)$$

**Where:**
- $N_{\text{correct}}$ is the number of instances where the predicted output exactly matches the reference.
- $N_{\text{total}}$ is the total number of instances evaluated.

- **Character Error Rate (CER)** measures the dissimilarity between the predicted and reference strings at the character level. It is calculated as the normalized Levenshtein distance between the two strings (see Equation (2)).

$$\text{CER} = \frac{\text{LevenshteinDistance}(w_p, w_r)}{len(w_r)} \quad (2)$$

**Where:**
- $w_p$ is the predicted native word,
- $w_r$ is the reference (actual) native word.
- $len(w_r)$ is the length of the reference (actual) native word.

The adopted methodology allowed us to thoroughly test specialized and general-purpose LLMs for transliteration of Romanized script into native Indian languages. The detailed steps of the overall methodology are described in Algorithm 1.

**Algorithm 1** Transliteration Testing Workflow

**Require:** Test dataset $\mathcal{D} = \{(x_i, y_i, L_i)\}_{i=1}^{N}$, where $x_i$ is the Romanized input, $y_i$ is the native-script word, and $L_i$ is the language
**Require:** Model $M$ (IndicXlit / General LLM / Fine-tuned LLM)
**Require:** Metric functions: `Top1Accuracy()`, `CER()`
**Ensure:** Performance scores over test set
1: Initialize: $N_{\text{correct}} \leftarrow 0$, $CER_{\text{total}} \leftarrow 0$
2: **for** each $(x_i, y_i, L_i)$ in $\mathcal{D}$ **do**
3: $\quad y_i^{\text{pred}} \leftarrow M.\text{predict}(x_i, L_i)$
4: $\quad$ **if** $y_i^{\text{pred}} == y_i$ **then**
5: $\quad\quad N_{\text{correct}} \leftarrow N_{\text{correct}} + 1$
6: $\quad$ **end if**
7: $\quad CER_{\text{total}} \leftarrow CER_{\text{total}} + \texttt{CER}(y_i, y_i^{\text{pred}})$
8: **end for**
9: Top-1 Accuracy $\leftarrow \frac{N_{\text{correct}}}{N} \times 100$
10: Average CER $\leftarrow \frac{CER_{\text{total}}}{N}$
11: **return** Top-1 Accuracy, Average CER

## 3 Results and Discussion

The findings indicate that general-purpose LLMs are, in fact, good transliterators and surpass IndicXlit for most of the instances. Detailed discussion of the results is presented next.

### 3.1 LLMs vs IndicXlit

Our findings indicate that LLMs belonging to GPT-family such as GPT-4o, GPT-4.5 and GPT-4.1 have show remarkable performance on almost all the Indian languages, for all the sub categories of the datasets, when compared to IndicXlit model. For example, GPT-4o got 62.54% accuracy and 0.129 CER value as compared to 55.26% accuracy and 0.141 CER value of IndicXlit model for Bengali langauge (Dakshina dataset). Similarly, GPT-4.5 (Top-1 Accuracy = 78.48% and CER = 0.041) outperformed IndicXlit (Top-1 Accuracy = 69.75% and CER = 0.052) for Tamil langauge when tested for AK-Freq subset. Similar results were observed for other langauges also such as Gujarati (AK-NEF subset) and Urdu (AK-NEI subset). The overall results are shown in Table 3.

Despite being outperformed by some of the LLMs, IndicXlit performed better than the other general purpose LLMs like Gemma-3 and Mistral-Large. For example, when tested for Kannada langauge instances (taken from Dakshina), IndicXlit demonstrated an accuracy of 76.90% and CER value of 0.056. On the other hand, Gemma-3 got 48.51% accuracy and 0.120 CER value while Mistral-Large got 46.20% accuracy and 0.124 CER value. However, for some instances (such as AK-NEF and AK-NEI datasets) Gemma-3 and Mistral-Large performed close to IndicXlit (for Hindi (AK-NEF), Gemma-3 accuracy = 55.57% and IndicXlit accuracy = 55.32%. For Urdu (AK-NEI), Mistral-Large accuracy = 44.97% and IndicXlit accuracy = 47.95%).

When considering only the general purpose LLMs, we found that GPT-4.5 performed the best followed by the other GPT models (GPT-4o and GPT-4.1 showed almost similar performance). Gemma-3 and Mistral-Large exhibited similar performance and were least accurate. With remarkable performances, general-purpose LLMs (especially GPT family) showed potential and capablity for Indic language transliteration.

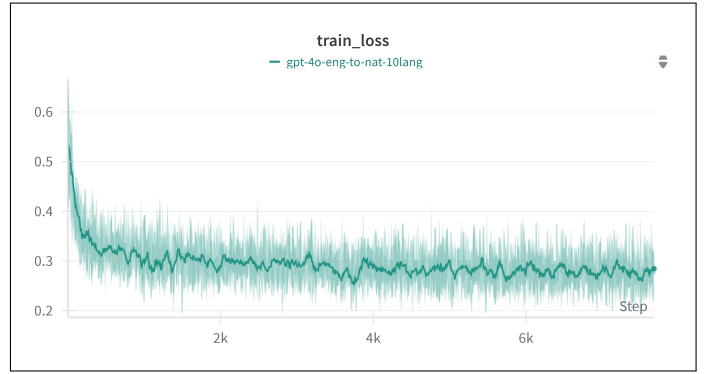

Figure 3: Training loss curve (with moving average smoothing) showing convergence behavior of GPT-4o during fine-tuning on a 10-language transliteration dataset.

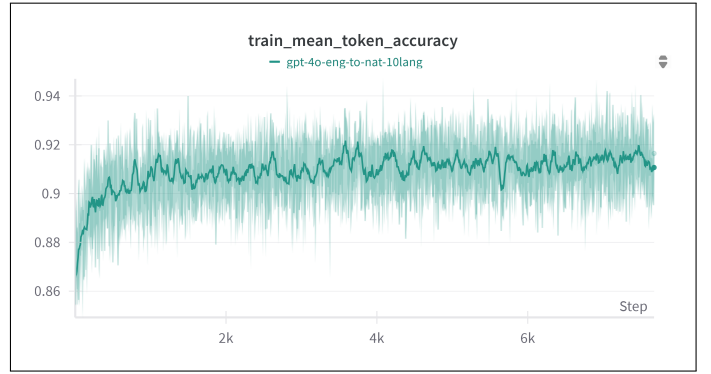

Figure 4: Mean token-level accuracy (with moving average smoothing) observed during GPT-4o fine-tuning.

### 3.2 Fine-tuned LLM vs IndicXlit

In our experiments, we fine-tuned GPT-4o on the Dakshina and Aksharantar datasets, and the results obtained are presented in Table 3.

The training loss for the fine-tuned GPT-4o model resulted in smooth convergence. The monotonous decreasing and flattening trend does not suggest signs of overfitting. The final recorded training loss was 0.27, and the corresponding validation loss was 0.24. The trajectory of the training curve over steps is shown in Figure 3. Figure 4 represents the mean token-level accuracy and shows its steady increment and stabilization around 91% throughout the training.

The results revealed that GPT-4o-Fine-Tuned model consistently outperformed IndicXlit in almost all languages when results are averaged across all dataset categories. It excels particularly in named entities datasets. For example, in the case of Hindi (AK-NEI dataset), GPT-4o-Fine-Tuned achieved 71.10% accuracy and 0.072 CER while IndicXlit got only 60.94% accuracy and 0.103 CER value. However, we observe that in specific cases such as for AK-Freq subcategory of Telugu language dataset IndicXlit performs better than GPT-4o-Fine-Tuned by 7% and shows almost 22% fewer character-level errors and for Kannada language IndicXlit outperformed GPT-4o-Fine-Tuned by minute difference of 0.1% in accuracy but GPT-4o-Fine-Tuned still gives fewer character-level errors compared to IndicXlit. Overall, while IndicXlit performs well primarily on the AK-Freq category, the GPT-4o-Fine-Tuned model demonstrated significantly stronger overall performance and achieved higher accuracy in 9 out of 10 languages. The average Top-1 accuracy of all models across all four datasets is pictorially shown in Figure 5.

| Models | | GPT-4o | | GPT-4.5 | | GPT-4.1 | | Gemma-3 | | Mistral-Large | | IndicXlit | | GPT-4o-Fine-Tuned | |
|---|---|---|---|---|---|---|---|---|---|---|---|---|---|---|---|
| Language | Dataset | Top-1 | Avg CER | Top-1 | Avg CER | Top-1 | Avg CER | Top-1 | Avg CER | Top-1 | Avg CER | Top-1 | Avg CER | Top-1 | Avg CER |
| Bengali | Dakshina | 62.54 | 0.129 | 66.97 | 0.119 | 60.65 | 0.135 | 36.79 | 0.226 | 35.22 | 0.222 | 55.26 | 0.141 | 68.10 | 0.108 |
| | AK-Freq | 64.42 | **0.092** | 65.70 | 0.097 | 60.67 | 0.104 | 39.77 | 0.171 | 46.42 | 0.157 | 61.58 | 0.107 | **69.43** | 0.095 |
| | AK-NEF | 37.41 | 0.233 | **37.65** | **0.232** | 37.16 | 0.235 | 35.57 | 0.256 | 27.14 | 0.290 | 33.20 | 0.241 | 36.68 | **0.232** |
| | AK-NEI | 44.88 | **0.171** | 45.30 | 0.174 | 44.79 | 0.176 | 40.47 | 0.204 | 33.98 | 0.216 | 35.08 | 0.203 | 42.94 | 0.178 |
| | Average | 52.31 | 0.156 | 53.90 | 0.156 | 50.82 | 0.162 | 38.15 | 0.214 | 35.69 | 0.221 | 46.28 | 0.173 | **54.29** | **0.153** |
| Gujarati | Dakshina | 56.37 | 0.124 | **64.51** | **0.102** | 54.91 | 0.131 | 35.85 | 0.185 | 28.55 | 0.219 | 61.26 | 0.106 | 61.39 | 0.109 |
| | AK-Freq | 51.29 | 0.106 | **65.78** | 0.084 | 49.46 | 0.115 | 29.54 | 0.175 | 34.13 | 0.165 | 61.53 | 0.094 | 65.42 | **0.078** |
| | AK-NEF | 46.70 | 0.161 | **49.27** | **0.142** | 47.56 | 0.153 | 45.97 | 0.152 | 34.60 | 0.242 | 41.45 | 0.178 | 46.78 | 0.152 |
| | AK-NEI | 51.48 | 0.131 | **52.08** | 0.129 | 49.87 | 0.140 | 47.24 | 0.146 | 35.82 | 0.188 | 45.37 | 0.150 | 50.32 | 0.136 |
| | Average | 51.46 | 0.130 | **57.91** | 0.114 | 50.45 | 0.135 | 39.65 | 0.164 | 33.28 | 0.204 | 52.40 | 0.132 | 55.98 | 0.119 |
| Hindi | Dakshina | 69.15 | 0.093 | 70.72 | 0.088 | 67.26 | 0.098 | 52.12 | 0.149 | 50.69 | 0.149 | 60.07 | 0.121 | **71.02** | **0.087** |
| | AK-Freq | 50.65 | 0.112 | 55.26 | 0.104 | 49.88 | 0.116 | 36.10 | 0.157 | 43.81 | 0.138 | 55.60 | 0.106 | **56.24** | **0.098** |
| | AK-NEF | 60.10 | 0.109 | 58.26 | 0.121 | 58.87 | 0.116 | 55.57 | 0.126 | 46.14 | 0.155 | 55.32 | 0.132 | **60.95** | **0.104** |
| | AK-NEI | 70.42 | 0.074 | **73.05** | **0.068** | 70.85 | 0.070 | 61.02 | 0.102 | 55.00 | 0.121 | 60.94 | 0.103 | 71.10 | 0.072 |
| | Average | 62.58 | 0.097 | 64.32 | 0.095 | 61.72 | 0.100 | 51.20 | 0.134 | 48.91 | 0.141 | 57.98 | 0.116 | **64.83** | **0.090** |
| Kannada | Dakshina | 74.73 | 0.056 | **79.85** | **0.048** | 73.40 | 0.060 | 48.51 | 0.120 | 46.20 | 0.124 | 76.90 | 0.056 | 76.82 | 0.051 |
| | AK-Freq | 64.20 | 0.065 | 72.55 | 0.051 | 62.75 | 0.068 | 26.05 | 0.157 | 33.13 | 0.121 | **76.16** | **0.045** | 69.76 | 0.055 |
| | AK-NEF | 45.15 | 0.152 | 47.24 | 0.148 | 43.56 | 0.157 | 44.54 | 0.155 | 26.50 | 0.229 | 51.36 | 0.135 | **53.67** | **0.132** |
| | AK-NEI | 49.96 | 0.125 | 50.21 | 0.127 | 50.21 | 0.125 | 42.91 | 0.157 | 32.29 | 0.198 | 47.25 | 0.137 | **51.10** | 0.121 |
| | Average | 58.51 | 0.099 | 62.46 | 0.094 | 57.48 | 0.102 | 40.50 | 0.147 | 34.53 | 0.168 | **62.92** | 0.093 | 62.84 | **0.089** |
| Malayalam | Dakshina | 59.69 | 0.128 | **71.61** | 0.095 | 60.29 | 0.124 | 43.01 | 0.184 | 25.22 | 0.299 | 63.35 | 0.108 | 70.20 | **0.086** |
| | AK-Freq | 46.00 | 0.109 | 60.75 | **0.086** | 48.72 | 0.105 | 23.18 | 0.179 | 16.19 | 0.242 | **64.65** | **0.065** | 62.59 | 0.071 |
| | AK-NEF | 31.49 | 0.250 | **32.10** | **0.239** | 31.12 | 0.252 | 27.92 | 0.267 | 11.79 | 0.434 | 28.81 | 0.268 | 31.24 | 0.254 |
| | AK-NEI | 36.87 | 0.223 | **39.08** | 0.206 | 36.53 | 0.224 | 32.37 | 0.245 | 18.03 | 0.367 | 36.69 | 0.228 | 37.45 | 0.216 |
| | Average | 43.51 | 0.178 | **50.88** | 0.156 | 44.16 | 0.176 | 31.62 | 0.219 | 17.81 | 0.336 | 48.38 | 0.167 | 50.37 | **0.157** |
| Marathi | Dakshina | 67.98 | 0.097 | **74.51** | **0.081** | 67.03 | 0.101 | 47.25 | 0.159 | 40.20 | 0.176 | 64.69 | 0.107 | 72.58 | 0.083 |
| | AK-Freq | 56.60 | 0.100 | **65.71** | 0.076 | 54.29 | 0.108 | 33.68 | 0.174 | 31.01 | 0.181 | 63.62 | 0.082 | 62.59 | **0.083** |
| | AK-NEF | 53.12 | 0.142 | 53.00 | 0.149 | 51.29 | 0.152 | 48.84 | 0.168 | 33.87 | 0.209 | 49.09 | 0.162 | **54.35** | **0.139** |
| | AK-NEI | 59.27 | 0.109 | **61.56** | **0.105** | 58.76 | 0.113 | 52.16 | 0.138 | 44.03 | 0.163 | 55.25 | 0.126 | 60.37 | 0.106 |
| | Average | 59.24 | 0.112 | **63.69** | **0.103** | 57.84 | 0.118 | 45.48 | 0.159 | 37.28 | 0.182 | 58.16 | 0.119 | 62.47 | **0.103** |
| Panjabi | Dakshina | 49.56 | 0.180 | **52.79** | 0.164 | 47.68 | 0.179 | 36.88 | 0.227 | 30.39 | 0.256 | 46.90 | 0.182 | 51.55 | 0.161 |
| | AK-Freq | 34.48 | 0.188 | 38.37 | 0.177 | 34.16 | 0.185 | 24.82 | 0.243 | 21.47 | 0.269 | **40.20** | **0.169** | 37.60 | 0.171 |
| | AK-NEF | 33.25 | 0.206 | 31.05 | 0.234 | 29.71 | 0.231 | 29.71 | 0.221 | 20.80 | 0.355 | 30.38 | 0.233 | **33.50** | **0.204** |
| | AK-NEI | 40.20 | 0.178 | **41.31** | 0.182 | 39.69 | 0.181 | 36.05 | 0.194 | 28.41 | 0.232 | 38.16 | 0.195 | 40.80 | **0.176** |
| | Average | 39.37 | 0.188 | **40.88** | 0.189 | 37.81 | 0.194 | 31.86 | 0.221 | 25.27 | 0.278 | 38.91 | 0.195 | 40.86 | **0.178** |
| Tamil | Dakshina | 70.44 | 0.108 | **76.61** | **0.096** | 67.91 | 0.116 | 46.80 | 0.173 | 49.28 | 0.178 | 67.97 | 0.117 | 72.84 | 0.104 |
| | AK-Freq | 74.07 | 0.046 | **78.48** | **0.041** | 72.54 | 0.050 | 43.77 | 0.114 | 51.46 | 0.094 | 69.75 | 0.052 | 75.22 | 0.041 |
| | AK-NEF | 44.74 | **0.134** | 43.52 | 0.138 | 40.22 | 0.151 | 42.91 | 0.141 | 24.03 | 0.268 | 39.08 | 0.163 | **45.11** | 0.137 |
| | AK-NEI | 48.94 | 0.123 | **49.36** | 0.122 | 48.60 | 0.127 | 45.12 | 0.144 | 27.71 | 0.216 | 43.58 | 0.147 | 48.60 | 0.124 |
| | Average | 59.55 | 0.103 | **61.99** | 0.099 | 57.32 | 0.111 | 44.65 | 0.143 | 38.12 | 0.189 | 55.09 | 0.119 | 60.44 | **0.102** |
| Telugu | Dakshina | 75.86 | 0.061 | **80.88** | **0.053** | 72.90 | 0.068 | 55.92 | 0.133 | 43.19 | 0.142 | 73.25 | 0.071 | 77.33 | 0.057 |
| | AK-Freq | 72.32 | 0.049 | 78.48 | 0.041 | 69.24 | 0.057 | 53.30 | 0.087 | 36.34 | 0.091 | **84.65** | **0.032** | 77.50 | 0.040 |
| | AK-NEF | 48.66 | 0.130 | 50.98 | 0.119 | 47.68 | 0.134 | 48.29 | 0.129 | 32.68 | 0.223 | 50.00 | 0.134 | **54.52** | **0.118** |
| | AK-NEI | 52.50 | 0.121 | 53.43 | 0.118 | 50.72 | 0.126 | 46.49 | 0.146 | 31.92 | 0.160 | 48.55 | 0.145 | **54.95** | **0.117** |
| | Average | 62.34 | 0.090 | 65.94 | **0.083** | 60.14 | 0.096 | 51.00 | 0.124 | 36.03 | 0.160 | 64.11 | 0.096 | **66.08** | **0.083** |
| Urdu | Dakshina | 49.96 | 0.177 | 54.19 | 0.165 | 48.40 | 0.184 | 36.36 | 0.244 | 36.28 | 0.232 | 41.56 | 0.200 | 51.31 | 0.168 |
| | AK-Freq | - | - | - | - | - | - | - | - | - | - | - | - | - | - |
| | AK-NEF | 57.35 | 0.114 | **62.87** | **0.094** | 55.39 | 0.119 | 52.82 | 0.126 | 46.75 | 0.188 | 48.65 | 0.151 | 57.23 | 0.113 |
| | AK-NEI | 53.58 | 0.112 | **56.56** | **0.101** | 53.75 | 0.115 | 39.95 | 0.165 | 44.97 | 0.156 | 47.95 | 0.146 | 55.79 | 0.111 |
| | Average | 53.63 | 0.134 | **57.87** | 0.120 | 52.51 | 0.139 | 43.04 | 0.178 | 42.67 | 0.192 | 46.05 | 0.166 | 54.78 | **0.131** |

Table 3: Table for comparison of Top-1 Accuracy and Character Error Rate (CER) of different Models

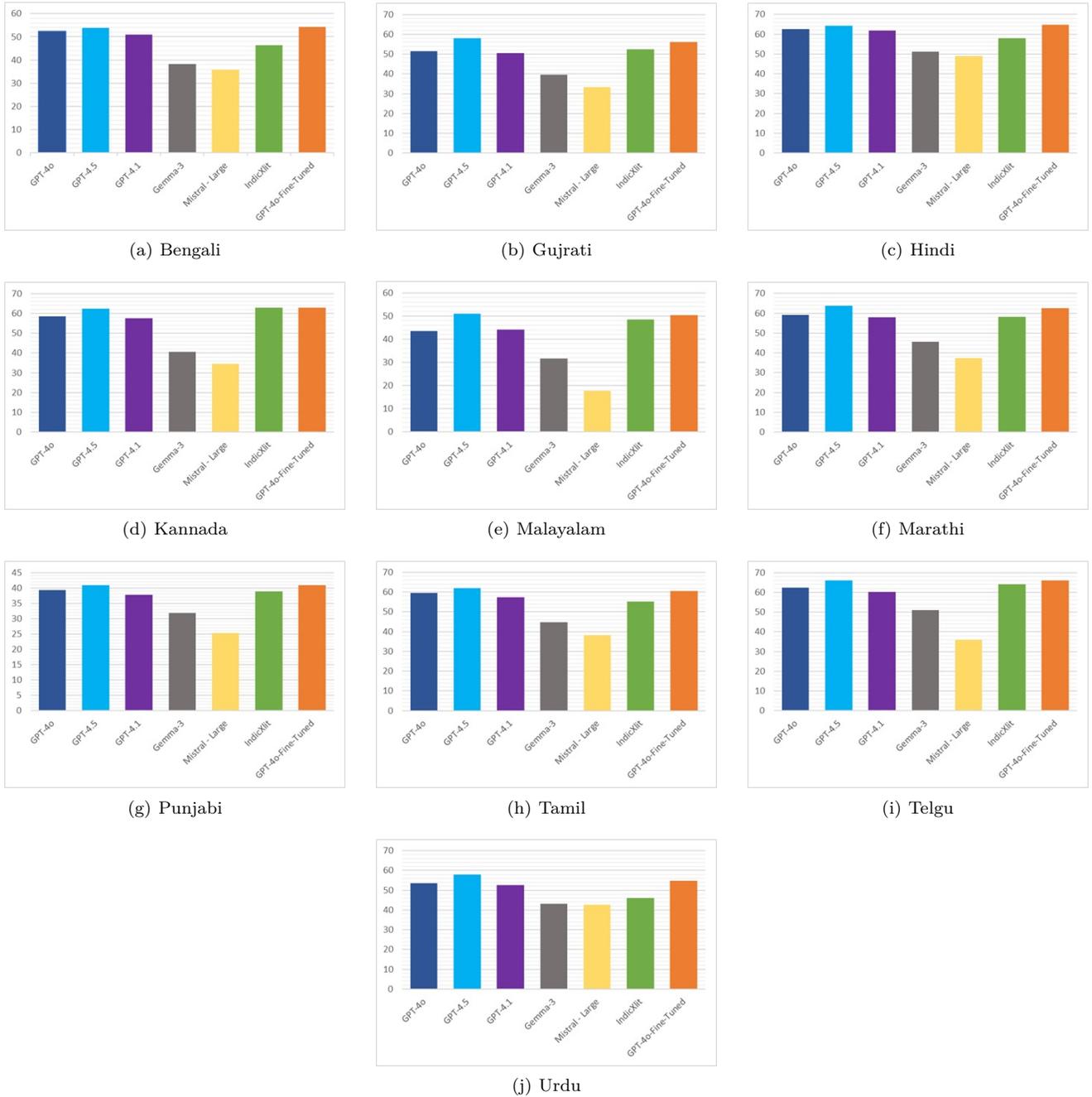

Figure 5: The average Top-1 accuracy of all models across all four datasets

### 3.3 Error Analysis

Apart from testing and fine-tuning of the models, we performed error analysis and robustness check for some of the selected best performing models. For error analysis, we utilized a small test set of around 1,000 word pairs, containing 100 word pairs from each of the ten selected languages spanning over all the four test set categories (keeping a balance of almost 25 word pairs per category). The purpose of error analysis was to understand the nature of errors in transliteration that occur commonly by each model.

We categorized the observed errors into three types: graphemic, script, and semantic. Graphemic errors occur when the output spelling is incorrect or inconsistent, despite the pronunciation being accurate. For example, for some instances of the Hindi language, GPT-4o produced **"आवाज़ों"** instead of **"आवाज़ों"**, GPT-4.5 produced **"ज्यादा"** instead of **"ज़्यादा"**, GPT-4o-Fine-Tuned produced **"सेलेक्ट"** instead of **"सलेक्ट"**, and IndicXlit returned **"टेनिस"** instead of **"टैनिस"**. The outputs reflect correct pronunciation but diverge from standard spelling conventions.

Script errors are more pronounced and affect the pronunciation due to incorrect character substitutions or insertions. Examples include GPT-4o generating **"कलापी"** for **"कालपी"**, GPT-4.5 giving **"अलांस"** instead of **"अलायंस"**, GPT-4o-Fine-Tuned outputting **"भोला"** instead of **"भोलू"**, and IndicXlit returning **"खायल"** for **"क़ायल"**. Semantic errors occur when the model generates a translation rather than a transliteration. While rare, we found instances in Marathi where GPT-4o and GPT-4o-Fine-Tuned produced **"आशियाई"** instead of **"एसेन"** for the Romanized word "asieen". Apart from these errors, few in-stances of other types were also observed. For example in case of Gujarati, GPT-4o generated "ફayetteville" instead of "ફેયતવિલ", and GPT-4o generated "భుజించవలENU" for Telugu langauge instead of "భుజించవలెను". Here, the predicted output mixes native script and Roman scripts.

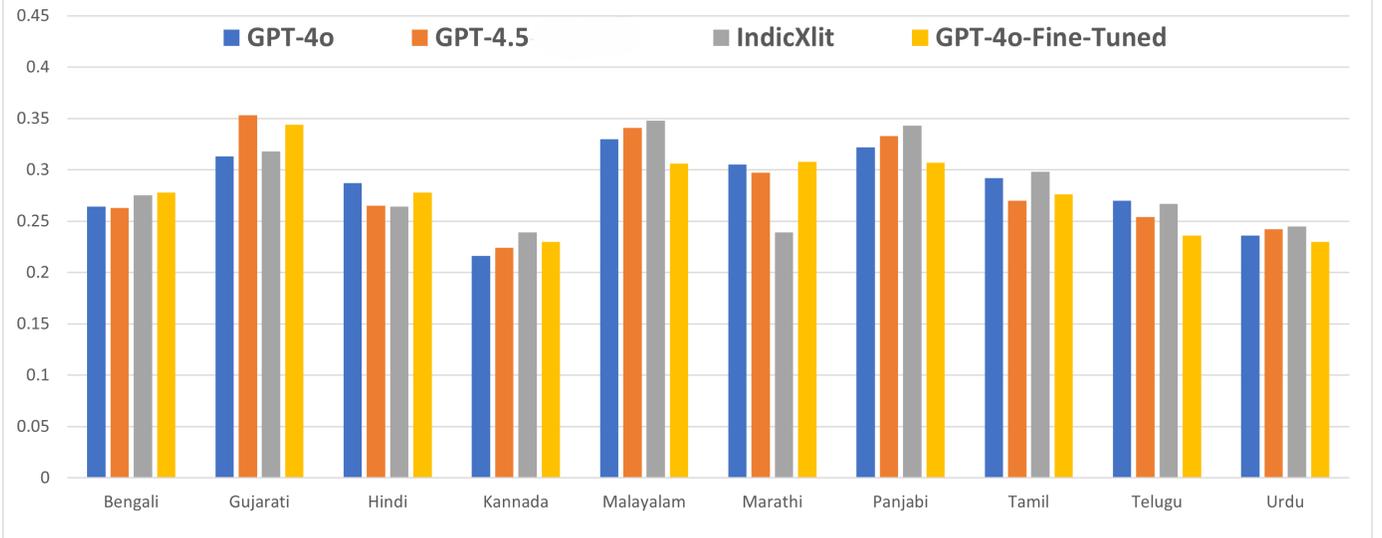

Figure 6: Bar graph illustrating mean CER values across models for each language. Lower bars indicate higher transliteration accuracy.

The comparative analysis of models reveals notable patterns in error distribution. Among the models in the GPT family, GPT-4.5 exhibits a tendency to produce a higher number of graphemic errors. However, it compensated with comparatively fewer script errors, which are more detrimental as they impact the phonetic integrity of the transliteration. In contrast, IndicXlit demonstrated a higher rate of script errors across multiple Indian languages. Across all models, semantic and other errors were rare. Fine-tuning GPT-4o also kept overall CER lower in four out of ten languages in comparison to other models (see Figure 6). For two languages, specialized IndicXlit model exhibited better performance and made fewer mistakes (average CER 0.239 in Marathi and 0.264 in Gujarati). Refer to Figure 7 for detailed error distribution.

### 3.4 Robustness

The inclusion of reasonable noise in the data enabled us to test how well these models work under real-world situations where inputs are not always clean or properly structured. We curated a test dataset of 100 common words from each of the ten languages. Since the considered dataset was relatively small, we manually added sensible noise to the Romanized script words in the following two ways:

- **Spelling based error**: Here we have slightly changed the spelling of the original words, but tried to keep almost the same pronunciation as that of the original word.
- **Case based error**: We added upper-case letters between the original words, still maintaining almost the same pronunciations as the original words.

Here are a few examples:

"nadiad" – "nadeead"
"texas" – "Teksas"
"lowell" – "loveLl"
"chaurasiya" – "chaurassia"

The results (see Table 4) indicated that the GPT-4o-Fine-Tuned model consistently achieved the highest transliteration accuracy and the lowest CER value across nearly all languages which demonstrates its relatively better generalization capabilities under noisy input. It is followed by GPT-4.5 and GPT-4o, which also perform well but with slightly higher error rates and reduced accuracies. In contrast, IndicXlit, despite being a model specifically designed for transliteration, exhibited lower accuracy scores and higher CER values. The contrast highlights the advantage of fine-tuning LLMs over specialized models for domain-specific tasks.

## 4 Limitations

While our study provides an exhaustive assessment of general-purpose LLMs and a task-specific transliteration model, some limitations are notable. The experiments were conducted on ten Indian languages. Though the major languages are covered, other languages can be included in future studies for better generalization of obtained results. Despite GPT-4.5 having returned the best results, fine-tuning could not be performed due to the unavailability of fine-tuning access on it [32]. Our analysis was limited to single-word transliteration. Sentence-level transliteration and context-sensitive mapping may result in different comparative outcomes and should be carried out in future. Moreover, the scope of the study was limited to transliteration from Romanized English to Indic native scripts. The reverse direction transliteration was not included in the current work.

## 5 Conclusion

Our work systematically compared general-purpose LLMs and a task-specific transliteration system, IndicXlit, over ten Indian languages on the task of transliterating rominaized script into their corresponding native script words. The outcomes proved that LLMs like GPT-4o, GPT-4.5, GPT-4.1, and fine-tuned GPT-4o model were able to rival or even surpass IndicXlit in most situations. Also, Fine-tuning GPT-4o improved transliteration accuracy and enabled it to beat IndicXlit for most scenarios.

The findings highlight a broader trend: LLMs, once considered generalists, are rapidly evolving into effective tools for domain-specific tasks. Their ability to adapt, generalize, and outperform bespoke systems signifies paradigm shift in AI where a few foundational models (or appropriately tuned

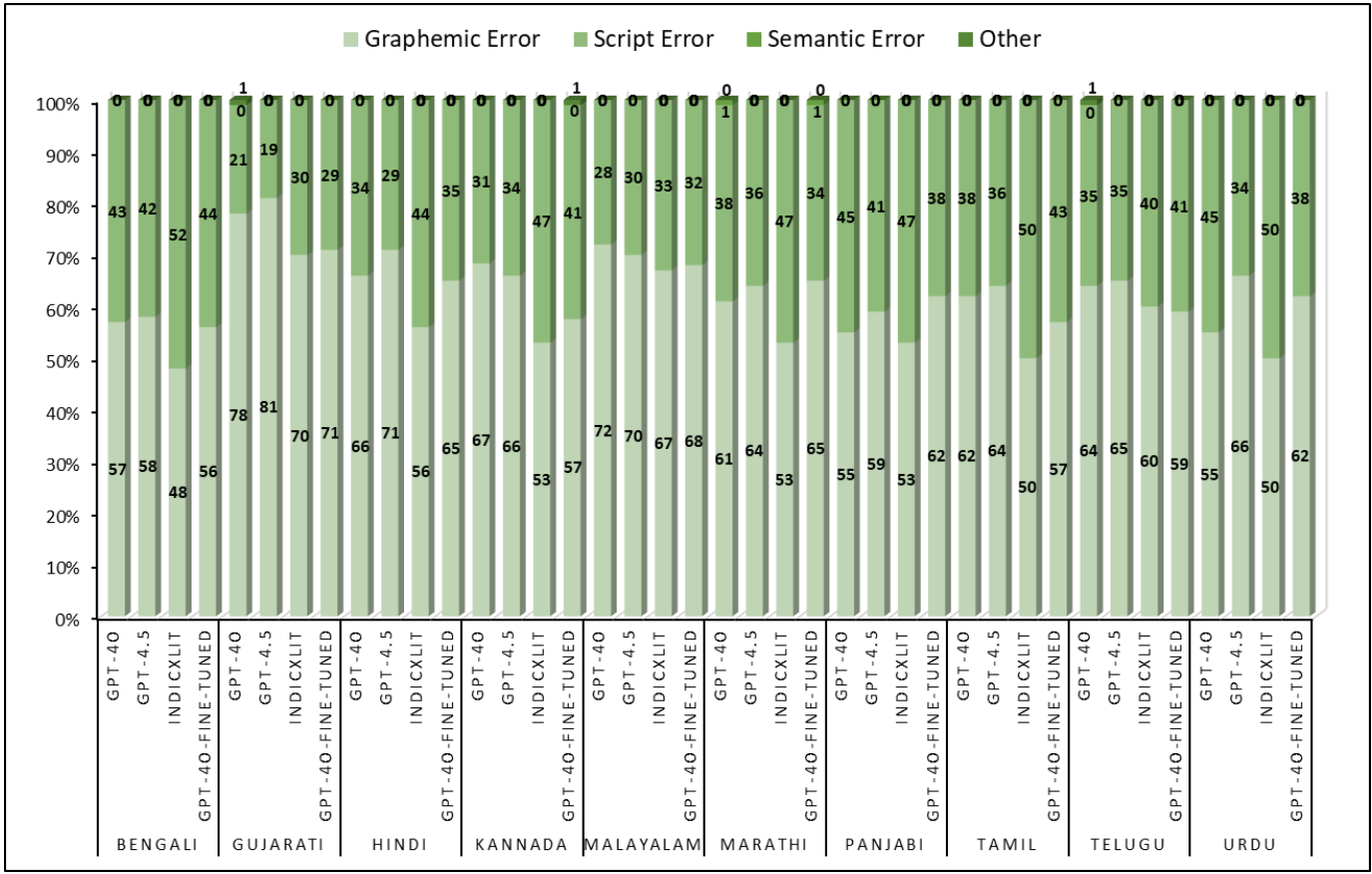

Figure 7: Distribution of error across different models and languages.

| Language | Models | | | | | | | |
|---|---|---|---|---|---|---|---|---|
| | GPT-4o | | GPT-4.5 | | IndicXlit | | GPT-4o-Fine-Tuned | |
| | Top-1 | Avg CER | Top-1 | Avg CER | Top-1 | Avg CER | Top-1 | Avg CER |
| Bengali | 19 | 0.318 | 21 | 0.339 | 18 | 0.331 | 27 | 0.282 |
| Gujarati | 24 | 0.266 | 29 | 0.234 | 17 | 0.264 | 29 | 0.217 |
| Hindi | 30 | 0.208 | 35 | 0.197 | 12 | 0.275 | 41 | 0.198 |
| Kannada | 15 | 0.264 | 22 | 0.232 | 17 | 0.217 | 26 | 0.241 |
| Malayalam | 16 | 0.318 | 23 | 0.328 | 12 | 0.341 | 35 | 0.261 |
| Marathi | 26 | 0.253 | 29 | 0.249 | 18 | 0.263 | 43 | 0.191 |
| Panjabi | 19 | 0.287 | 19 | 0.294 | 10 | 0.315 | 25 | 0.251 |
| Tamil | 19 | 0.218 | 22 | 0.218 | 20 | 0.232 | 37 | 0.172 |
| Telugu | 24 | 0.235 | 26 | 0.213 | 20 | 0.225 | 35 | 0.205 |
| Urdu | 36 | 0.236 | 35 | 0.221 | 26 | 0.272 | 44 | 0.201 |

Table 4: Performance Comparison of Models Across Languages with Noisy Inputs

LLMs) can serve a wide spectrum of specialized applications with minimal overhead. Such convergence of generality and specialization holds transformative potential for resource-constrained settings and multilingual domains. Aspects of future study could include language extension covering more Indian languages, assessment of sentence-level transliteration across languages, investigation into fine-tuning of more recent models and testing LLMs for other domain-specific tasks.